\documentclass{article}
\usepackage{iclr2027_conference,times}

\usepackage{amsmath,amsfonts,bm}

\def\eqref#1{equation~\ref{#1}}

\def\1{\bm{1}}

\DeclareMathAlphabet{\mathsfit}{\encodingdefault}{\sfdefault}{m}{sl}
\SetMathAlphabet{\mathsfit}{bold}{\encodingdefault}{\sfdefault}{bx}{n}

\usepackage{amsmath,amssymb,amsthm,mathtools}
\usepackage{booktabs}
\usepackage{graphicx}
\usepackage{microtype}
\usepackage{hyperref}
\usepackage{url}
\usepackage{xcolor}
\usepackage{enumitem}

\newtheorem{theorem}{Theorem}
\newtheorem{corollary}{Corollary}
\newtheorem{proposition}{Proposition}

\theoremstyle{definition}
\newtheorem{definition}{Definition}
\theoremstyle{remark}
\newtheorem{remark}{Remark}

\newcommand{\A}{\mathcal{A}}
\newcommand{\ThetaSet}{\Theta}
\newcommand{\Risk}{R}
\newcommand{\TV}{\operatorname{TV}}
\newcommand{\keep}{\textsc{Keep }}
\newcommand{\recent}{\textsc{Recenter }}
\newcommand{\Normal}{\mathcal{N}}
\newcommand{\Ebb}{\mathbb{E}}
\newcommand{\Pbb}{\mathbb{P}}

\title{When Is Test-Time Adaptation Identifiable from Unlabeled Evidence?}

\author{Kartik Jhawar\thanks{Corresponding author: \texttt{kartikvi001@e.ntu.edu.sg}} \quad \& \quad Lipo Wang \\
Institute for Digital Molecular Analytics and Science\\
School of Electrical and Electronic Engineering\\
Nanyang Technological University, Singapore\\
\texttt{kartikvi001@e.ntu.edu.sg}
}

\iclrfinalcopy

\begin{document}
\maketitle

\begin{abstract}
Test-time adaptation (TTA) offers many ways to update a deployed model without labels, but choosing the wrong update can make a strong source model worse. Recent methods therefore try to predict which adaptation will work from unlabeled test data. We ask a prior question: \emph{does the evidence given to the selector contain enough information to determine the best action at all?} We show that this is not guaranteed, even with a perfect selector. If an observation channel makes two deployments look the same while their TTA rankings differ, reliable selection is impossible from that channel; richer evidence can restore the decision only when it resolves the relevant ambiguity. We make this boundary exact in a finite-batch Gaussian TTA model, where doing nothing beats mean recentering for small shifts, recentering wins beyond a unique critical shift, and the boundary shrinks as $1/\sqrt n$. Public benchmark studies on CIFAR-100-C and DomainNet-126 show the same failure mode with modern TTA methods: changing only deployment structure can reverse the oracle action while global order-blind evidence remains unchanged. The result is a practical way to separate two failure modes that are usually mixed together: a weak selector versus an information channel that cannot support the desired decision in the first place.
\end{abstract}

\section{Introduction}
Test-time adaptation (TTA) updates a trained model using unlabeled data after deployment. Methods such as Tent, EATA, SAR, DeYO, and ROID can improve performance when test data differ from the source distribution \citep{wang2021tent,niu2022eata,niu2023sar,lee2024deyo,marsden2024roid}. But the same update can also be harmful. Small batches, class imbalance, temporal correlation, and realistic model-selection choices can change or even reverse the benefit of adaptation \citep{gong2022note,lim2023ttn,zhao2023pitfalls,sreeram2026tempora}.

This has created a second problem on top of adaptation itself: \emph{which TTA procedure should we use?} AETTA and TTALine estimate TTA performance without target labels \citep{lee2024aetta,kim2024ttaline}. MORPHEUS goes closer to our setting and predicts which candidate TTA method will perform best from pre-adaptation entropy and representation geometry \citep{danilowski2026morpheus}. Cygert et al. study unsupervised TTA model selection after candidate adaptations have been executed \citep{cygert2026realistic}. Recent theory and diagnostics also show that TTA can be hard to recover, temporally unstable, or underspecified \citep{zhou2026learnability,sreeram2026tempora,shamsi2026multihypothesis}. These works improve the selector or the adaptation procedure. Also, we have stated works to tell us when TTA can become helpful or harmful, but none so far tell us
whether an unlabeled selector can recognize, from the evidence available before adaptation, which side
of that reversal it is on. We therefore ask a question that comes before both:

\begin{quote}
\emph{Is the unlabeled evidence available before adaptation actually enough to determine which action is best?}
\end{quote}

This question is not the obvious statement that the best method depends on the candidate family. We keep the candidate actions (our menu of TTA choices - TENT, DeYO, etc.) fixed. The problem is that the selector sees only a chosen summary of the unlabeled deployment data, while the relative performance of stateful TTA actions can depend on deployment properties---such as batch composition and order---that this summary may not preserve. Two deployments can therefore be indistinguishable through the selector's evidence and still have opposite action rankings. In that case, a more powerful regressor, a larger neural selector, or more training data for the selector cannot recover information that the observation channel discarded.

We call the oracle action \emph{identifiable} when the allowed evidence is sufficient to determine at least one oracle-best intervention over the deployment family under study. This is intentionally a decision-level question. We do not claim that identifying an optimal action without identifying all risks is a new statistical principle; partial-identification and decision theory already study optimal decisions under incomplete knowledge \citep{kasy2016partial,pu2021optimal,han2024dynamic,christensen2026optimal}. Domain-adaptation theory likewise gives important impossibility and identifiability precedents \citep{benDavid2010impossibility,garg2022atc,gulrajani2022identifiability,kong2022partial,hanneke2023limits,dong2026hardness}. Our contribution is to study TTA selection as an information problem: the actions themselves are data-dependent interventions, the observation channel is fixed before choosing an action, and we ask exactly how batch size, shift mechanism, stream structure, and evidence richness change the answer.

Our contribution is to study TTA selection as an information problem: before asking how to build a
better selector, we ask whether the unlabeled evidence available to it is sufficient to determine the
best action.

Our contributions are:
\begin{itemize}[leftmargin=1.2em,itemsep=2pt,topsep=3pt]

    \item \textbf{A formulation of TTA action identifiability.}
    We formalize when the oracle-best action from a fixed set of TTA choices is determined by a
    specified unlabeled observation channel and deployment family.

    \item \textbf{Information limits and positive conditions for selection.}
    We show that indistinguishable deployments with different best actions make reliable selection
    impossible from that evidence, and give the complementary condition under which an optimal
    action is identifiable.

    \item \textbf{A TTA-specific finite-batch boundary.}
    In a tractable KEEP-versus-RECENTER model, we prove a unique transition between when
    adaptation hurts and helps, show that its critical shift scales as $1/\sqrt{n}$, and construct
    different shift mechanisms with the same coarse evidence but opposite preferred actions.

    \item \textbf{Public-benchmark evidence of the same information boundary.}
    On CIFAR-100-C and DomainNet-126, the same selected images can require different TTA actions
    after only the deployment structure changes. Global order-blind evidence cannot capture these
    reversals, while stream-aware evidence can; under different stochastic deployments, this advantage
    can disappear.
\end{itemize}


The practical message is direct. Before asking how to build a stronger TTA selector, we should first ask whether its inputs can support the decision we expect it to make. This separates a modeling failure from an information failure, and tells us when adding richer deployment evidence is more important than adding selector capacity.

\section{Problem Setup}
\label{sec:setup}

\paragraph{Actions.}
Let $f_0$ be a fixed pretrained source model. We have a finite action set
\[
\A=\{\keep,a_1,\ldots,a_K\}.
\]
Here \keep means exactly one thing: \emph{use the frozen source model and do not adapt it}. Every other action may be a data-dependent TTA procedure. Let $X_{1:n} = (X_1, \dots, X_n)$ denote an unlabeled target adaptation batch of size $n$. Formally, an action can be viewed as a map
\[
\mathcal{M}_a:(f_0,X_{1:n})\mapsto f_{a,X_{1:n}},
\]
which takes the source model and an unlabeled adaptation batch and returns the predictor that will be deployed.

In the benchmark experiments, the five empirical actions are SOURCE/KEEP, DeYO, Tent, ROID, and test-time normalization \citep{lee2024deyo,wang2021tent,marsden2024roid}. To theoretically analyze the finite-batch trade-offs between estimation noise, shift magnitude, and shift mechanisms in closed form, Section~\ref{sec:gaussian_boundary} introduces \text{RECENTER} as an analytically tractable one-dimensional prototype of adaptation (shifting a decision threshold to the unlabeled sample mean). \text{RECENTER} is strictly a theoretical device to study \text{KEEP}-versus-adaptation boundaries and should not be confused with the empirical benchmark methods or with NEO \citep{murphy2026neo}.

\paragraph{A ``world''.}
A world $\theta\in\ThetaSet$ specifies both the target joint distribution $P_\theta(X,Y)$ and the deployment protocol, including batch construction, ordering, or temporal structure when these affect adaptation. This second part is essential because a stateful TTA action can produce a different predictor from the same image multiset when the images arrive differently.

We define the prospective risk of action $a$ on an independent fresh target point $(X',Y')$ as
\[
\Risk_\theta(a)
=\Ebb_\theta\!\left[
\ell\!\left(f_{a,X_{1:n}}(X'),Y'\right)
\right],
\]
where the expectation includes both the adaptation batch and the fresh evaluation point. The oracle action set is
\[
\A^*(\theta)=\arg\min_{a\in\A}\Risk_\theta(a).
\]
Target labels are used only after an experiment to measure these risks and define the oracle. A real selector never sees them before choosing an action.

\paragraph{What the selector is allowed to see.}
Before choosing an action, the selector observes only
\[
Z_n=\phi_n(f_0,X_1,\ldots,X_n),
\]
where $\phi_n$ denotes the specified unlabeled evidence channel, which instantiates features such as entropy, confidence margins, source logits, global empirical moments, or sequential order statistics. Let $Q_{\theta,n}$ be the probability law of $Z_n$ in world $\theta$.

\begin{definition}[Oracle-action identifiability]
Fix an observation channel and a family of worlds. For an evidence law $Q$, define
\[
\ThetaSet(Q)=\{\theta\in\ThetaSet:Q_{\theta,n}=Q\}.
\]
The oracle action is identifiable at $Q$ if
\[
\bigcap_{\theta\in\ThetaSet(Q)}\A^*(\theta)\neq\varnothing.
\]
If every world has one unique best action, this simply means: any two worlds that look the same through the allowed evidence must have the same best action.
\end{definition}

\paragraph{Why we define it this way.}
This definition asks an information question, not a training question. If we knew the full probability law of the allowed evidence, would it determine a best action? A practical selector sees only finite data, so there is a second statistical problem of learning the decision from limited samples. We keep these questions separate.

It is also useful to separate three levels of difficulty:
\[
\text{Target/Mechanism-ID}\Rightarrow\text{Risk-ID}\Rightarrow\text{Action-ID}.
\]
Knowing the complete target mechanism is stronger than knowing every action risk, and knowing every action risk is stronger than merely knowing which action wins. The reverse directions need not hold. This hierarchy is a positioning device rather than a claim of a new general decision-theory principle; Section~\ref{sec:related} connects it to earlier partial-identification and domain-adaptation work.

\paragraph{Regret.}
For a selector $s$, its oracle regret in world $\theta$ is the extra risk caused by its chosen action compared with the oracle:
\[
\operatorname{Reg}_\theta(s)
=\Ebb\!\left[\Risk_\theta(s(Z_n))-\min_{a\in\A}\Risk_\theta(a)\right].
\]
The theory below asks when this extra risk can be forced to zero and when some positive error is unavoidable.

\section{When Is the Oracle TTA Action Identifiable?}
\label{sec:theory}

\subsection{If two worlds look the same but need different actions, selection is impossible}

\begin{theorem}[Exact non-identifiability]
\label{thm:exact}
Suppose two worlds $\theta_0,\theta_1$ have exactly the same evidence law,
$Q_{\theta_0,n}=Q_{\theta_1,n}$, but have different unique oracle actions $a_0^*\neq a_1^*$. Under an equal prior over the two worlds, every selector based only on $Z_n$ has average oracle-action error at least $1/2$.

If every non-oracle action costs at least $\Delta>0$ extra risk, then every selector has average oracle regret at least $\Delta/2$.
\end{theorem}

\paragraph{Simple meaning.}
The selector sees the same kind of evidence in both worlds, so it has no reliable way to know which world it is in. If the correct actions disagree, it cannot be right in both. Under an equal two-world test, at least half of the error is unavoidable.

The statement can remain true even with unlimited unlabeled data if the full allowed unlabeled observation remains the same. This is the TTA specialization of a classical two-point decision/testing argument; we use it as a limit on pre-action selectors rather than claim the testing inequality itself as new. In our discrete construction, both worlds have the same unlabeled marginal $P(X)=(0.45,0.25,0.30)$, but the action risks reverse:

\begin{center}
\begin{tabular}{lcc}
\toprule
& \keep & \textsc{ADAPT}\\
\midrule
World A & 0.5625 & \textbf{0.2500}\\
World B & \textbf{0.2100} & 0.5275\\
\bottomrule
\end{tabular}
\end{center}

Perfect equality is a strong condition, so we also need an approximate statement.

\begin{corollary}[Approximate non-identifiability]
\label{cor:tv}
Let the two evidence laws be $Q_0,Q_1$ and let their unique oracle actions differ. Under an equal prior,
\[
\Pbb(\text{wrong oracle action})\geq
\frac{1-\TV(Q_0,Q_1)}{2}.
\]
If every wrong action costs at least $\Delta$, then
\[
\Ebb[\operatorname{Reg}]
\geq
\frac{\Delta}{2}\bigl(1-\TV(Q_0,Q_1)\bigr).
\]
\end{corollary}

\paragraph{Simple meaning.}
Total variation (TV) measures how distinguishable two evidence distributions are. TV near zero means they look very similar. The bound says that when the evidence distributions overlap heavily, some action-selection error must remain. Complete proofs of Theorem~\ref{thm:exact} and Corollary~\ref{cor:tv} are in Appendix~\ref{proof1} and \ref{proof2}, respectively.

\subsection{The positive side: when the action is identifiable}

\begin{theorem}[Compatibility criterion]
\label{thm:positive}
Fix an evidence law $Q$. There exists one action, depending only on $Q$, that has zero oracle regret for every world compatible with $Q$ if and only if
\[
\bigcap_{\theta\in\ThetaSet(Q)}\A^*(\theta)\neq\varnothing.
\]
If every compatible world has a unique oracle action, this means they must all share the same action.
\end{theorem}

\paragraph{Simple meaning.}
After seeing the evidence, imagine making a list of every hidden world that could still have produced it. If all of those worlds agree on at least one best action, then the action is determined even if the full target world is not.

The complete proof is in Appendix~\ref{proof3}. A stronger sufficient condition is a positive worst-case margin:
\[
M_a(Q)=\inf_{\theta\in\ThetaSet(Q)}\min_{b\neq a}
\left[\Risk_\theta(b)-\Risk_\theta(a)\right]>0.
\]
Then $a$ is not only optimal in every compatible world; it beats every other action by a positive amount.

\subsection{Why ignoring order can hide the correct action}

\begin{proposition}[Permutation/invariance obstruction]
\label{prop:invariance}
Let $g$ be a transformation such as permuting the order of a deployment batch. Suppose the evidence channel ignores that transformation, so $\phi_n(gB)=\phi_n(B)$ for every batch $B$. If a world and its transformed version have different unique oracle actions, then the oracle action is not identifiable from $\phi_n$ over any family containing both worlds.
\end{proposition}

\paragraph{Simple meaning.}
If the selector throws order away, it cannot react to order. But a stateful TTA method may react strongly to order. Therefore the same image set can look identical to the selector while requiring a different action. This is exactly what our matched CIFAR-100-C and DomainNet-126 experiments are designed to test. The formal proof is in Appendix~\ref{proof4}.

\subsection{A TTA-Specific Finite-Batch KEEP-versus-RECENTER Boundary}
\label{sec:gaussian_boundary}

The previous results are general information statements. We now want a concrete TTA mechanism where the best action changes in a predictable way with batch size and shift size. This is the paper's main TTA-specific theorem.

\paragraph{The Core Trade-off.} When adapting to an unlabeled batch, an algorithm faces a fundamental tension:
\begin{itemize}
    \item \textbf{\textsc{Keep} (Frozen Source):} Incurs \emph{zero} estimation noise, but suffers increasing error as the true distribution shift $|\delta|$ grows.
    \item \textbf{\textsc{Recenter} (Adaptation):} Successfully eliminates the true shift $\delta$, but incurs a \emph{finite-sample noise floor} because the shift is estimated from a noisy batch of size $n$.
\end{itemize}

\begin{figure}[t]
\centering
\includegraphics[width=\linewidth]{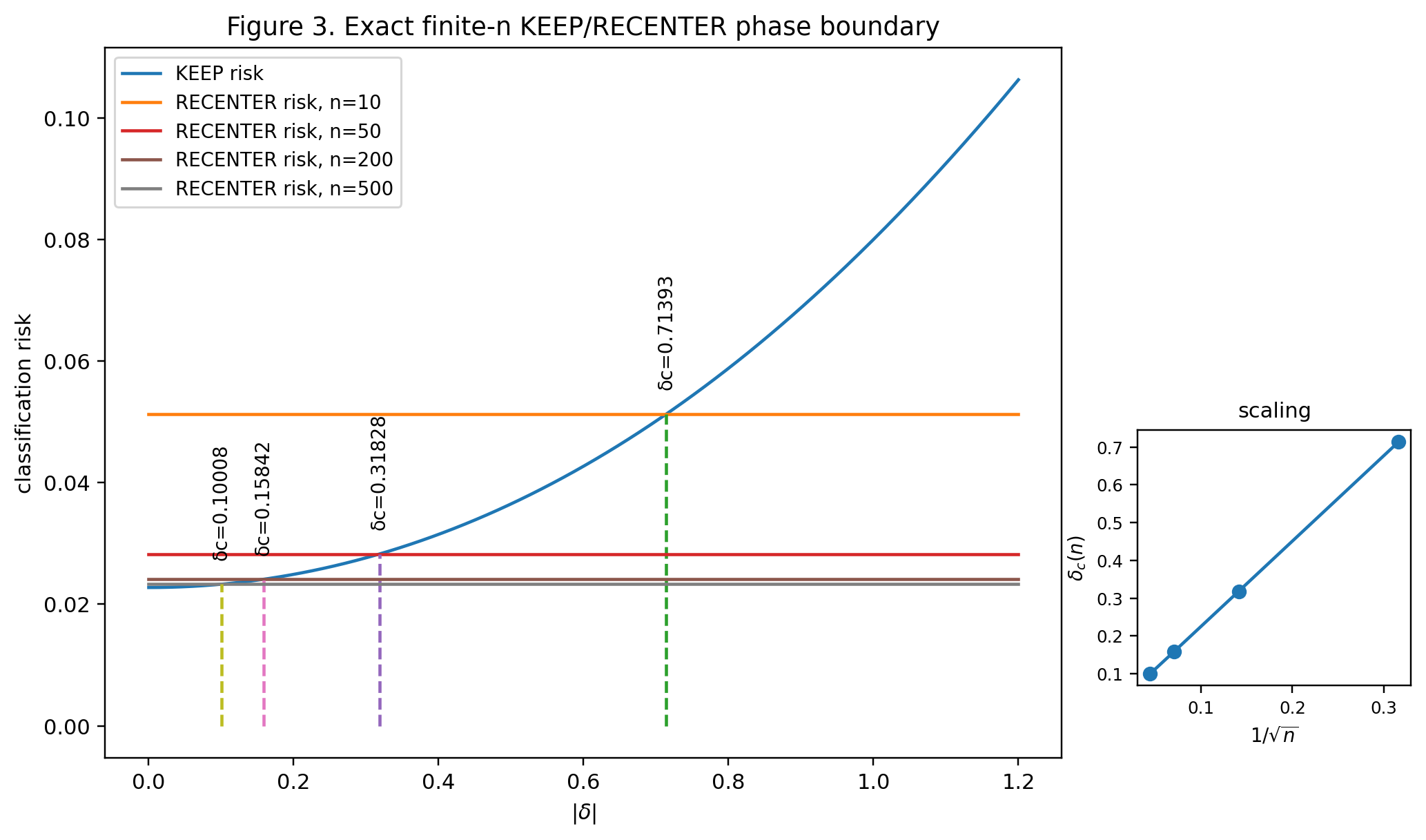}
\caption{\textbf{Finite-batch \textsc{Keep} vs.\ \textsc{Recenter} phase boundary (Theorem~\ref{thm:gaussian_boundary}).} 
\emph{Left:} Prospective classification risk under Gaussian translation shift ($\mu=2, \sigma=1$). \textsc{Keep} (blue) is optimal near zero shift, while \textsc{Recenter} (horizontal lines) wins once $|\delta| > \delta_c(n)$. 
\emph{Right:} Critical tipping point $\delta_c(n)$ plotted against $1/\sqrt{n}$, confirming the exact $\Theta(n^{-1/2})$ finite-batch rate.}
\label{fig:gaussian_boundary}
\end{figure}

\paragraph{1D Gaussian Setup.} Consider balanced binary classification $Y \in \{-1, +1\}$ with source distributions $\mathcal{N}(\pm\mu, \sigma^2)$ ($\mu, \sigma > 0$) split by a source threshold at zero. In the target domain, data undergo a translation shift $\delta$:
\begin{equation}
X = \delta + \mu Y + \epsilon, \quad \epsilon \sim \mathcal{N}(0, \sigma^2).
\end{equation}
\textsc{Keep} leaves the threshold at zero. \textsc{Recenter} receives $n$ unlabeled target samples $X_{1:n}$, computes the empirical sample mean $\hat{\delta} = \bar{X}_n$, and classifies a fresh target point using threshold $\hat{\delta}$.

The prospective risk of \textsc{Keep} grows strictly with the true translation $|\delta|$:
\begin{equation}
R_K(\delta) = \frac{1}{2}\left[\Phi\left(-\frac{\mu+\delta}{\sigma}\right) + \Phi\left(\frac{\delta-\mu}{\sigma}\right)\right].
\end{equation}
For \textsc{Recenter}, shifting the threshold cancels $\delta$, leaving only the finite-sample estimation error $E_n = \hat{\delta} - \delta$. With $S_n = \sum_{i=1}^n Y_i$, the exact finite-$n$ risk is independent of $\delta$ and given by:
\begin{equation}
\label{eq:recenter-risk}
R_R(n) = 2^{-n} \sum_{k=0}^n \binom{n}{k} \Phi\left( \frac{\mu(2k-n)/n - \mu}{\sigma \sqrt{1 + 1/n}} \right).
\end{equation}

\begin{theorem}[Finite-batch \textsc{Keep}/\textsc{Recenter} boundary]
\label{thm:gaussian_boundary}
For every finite batch size $n \ge 1$, there exists a unique critical shift $\delta_c(n) > 0$ such that $R_K(\delta_c(n)) = R_R(n)$. Furthermore:
\begin{align}
|\delta| < \delta_c(n) &\implies R_K(\delta) < R_R(n) \quad \text{(\textsc{Keep} wins)}, \\
|\delta| > \delta_c(n) &\implies R_K(\delta) > R_R(n) \quad \text{(\textsc{Recenter} wins)}.
\end{align}
As $n \to \infty$, this phase boundary scales as:
\begin{equation}
\delta_c(n) = \frac{\sqrt{\mu^2 + \sigma^2}}{\sqrt{n}} + O(n^{-3/2}) = \Theta(n^{-1/2}).
\end{equation}
\end{theorem}

\paragraph{Intuition and the $1/\sqrt{n}$ Law.} When shift $\delta = 0$, the source threshold is already optimal. \textsc{Recenter} still estimates a mean from finite samples, introducing random jitter that degrades performance ($R_R(n) > R_K(0)$). However, as $|\delta|$ grows, \textsc{Keep} degrades quadratically, while \textsc{Recenter} maintains a flat noise floor. The two curves must cross at exactly one critical value $\delta_c(n)$ (Figure~\ref{fig:gaussian_boundary}). 

Because the standard error of a sample mean scales as $1/\sqrt{n}$, the true distribution shift must exceed this statistical noise scale before test-time adaptation provides a net benefit. The full proof in Appendix~\ref{proof5} uses the exact finite mixture above; the numerical table does not rely on a normal approximation.

\subsection{The same target mean can come from two different shifts}
\label{sec:prior-shift}

A change in target mean does not always mean that the whole distribution translated. Suppose the class-conditionals stay fixed at $\Normal(\pm\mu,\sigma^2)$ but the positive-class probability changes to $\pi\neq1/2$. Then
\[
m=\mu(2\pi-1)
\]
is the new target mean even though neither class distribution moved.

\begin{proposition}[Under pure prior shift, mean recentering moves the threshold the wrong way]
\label{prop:prior}
Under pure class-prior shift with $\pi\neq1/2$, population RECENTER has strictly larger classification risk than KEEP.
\end{proposition}

\paragraph{Simple meaning.}
If there are simply more positive examples, the overall mean moves to the right. Mean recentering interprets this as a physical translation and moves the threshold right. But the Bayes-optimal threshold actually moves left because positive examples are now more common. The recentering correction therefore goes in the wrong direction. The Full Proof is in Appendix~\ref{proof6}.

\begin{corollary}[Same population mean, opposite best actions]
\label{cor:mean}
Fix $m$ with $0<|m|<\mu$. Compare two worlds:
\begin{enumerate}[leftmargin=*]
\item World C: balanced classes with a real translation $\delta=m$;
\item World P: no translation, but class prior $\pi=(1+m/\mu)/2$.
\end{enumerate}
Both worlds have the same population target mean $m$. Population RECENTER is better in World C, while KEEP is better in World P. Therefore a channel that reveals only the population target mean cannot identify the oracle action over a family containing both worlds.
\end{corollary}

\begin{remark}[What ``same mean'' does not mean]
The two worlds have the same \emph{population} mean. Their full distributions are not identical, and the probability law of a finite-sample empirical mean need not be identical either. This is intentional. The result shows that mean-only evidence is too coarse; richer statistics may separate the worlds and recover the correct action. The Full Proof is in Appendix~\ref{proof7}.
\end{remark}

Together, these results give the main theoretical message: the best TTA action depends not only on the shift itself, but also on batch size, the shift mechanism, the deployment structure, and what information the selector is allowed to use.

\section{Experimental Design}
\label{sec:methods}

The experiments are built to test the information question above. They are not presented as a new universal TTA algorithm.

\subsection{Candidate actions and oracle evaluation}
Both public-benchmark studies use the same five choices: SOURCE/KEEP, DeYO, Tent, ROID, and test-time normalization \citep{lee2024deyo,wang2021tent,marsden2024roid}. Every action starts from the same frozen source state for a target world. Labels are used only afterward to measure each action's error and define the oracle. The selector evidence is always label-free and computed before the selected TTA action is executed.

\subsection{CIFAR-100-C: controlled deployments}
CIFAR-100-C \citep{hendrycks2019benchmark} gives the broad controlled study. We construct 675 worlds: 15 corruptions, three severities, five deployment regimes, and three seeds. The regimes change class-prior conditions and stream structure. In the key matched experiment, an IID world and a correlated world contain exactly the same selected images; only their ordering changes.

This directly tests Proposition~\ref{prop:invariance}. A global evidence channel that ignores order cannot distinguish the two arrangements, while stateful TTA actions can behave differently on them.

We compare a hierarchy of source-only evidence. Z1 contains simple global uncertainty summaries. Z2 adds richer global, permutation-invariant source-output statistics. Z3 additionally retains how source-model behavior changes across successive windows. Z4 further adds a fixed projection of the ordered source logits. The selector is evaluated mainly by oracle regret, together with exact/tie-aware action accuracy, negative transfer, benefit capture, bootstrap intervals, and paired tests. Appendix~\ref{app:experiments} lists the complete reviewer-check suite.

\paragraph{Selection baselines.}
SOURCE/KEEP and the best fixed action measure how much can be achieved without world-specific selection. MORPHEUS is the closest direct pre-action comparison and is implemented using the author-confirmed MultiOutput random-forest architecture and both documented Neural-Collapse feature interpretations \citep{danilowski2026morpheus}. We also report the same regression architecture over our Z1--Z4 evidence hierarchy. Cygert et al. are discussed as a stronger-information-budget comparison because their unsupervised selection statistics are evaluated after candidate adaptations have been run \citep{cygert2026realistic}; we therefore do not pretend that it is a same-budget pre-action baseline.

\subsection{DomainNet-126: natural domain shifts}
DomainNet-126 \citep{peng2019moment} tests whether the same phenomenon survives beyond corruption shifts. We use 90 worlds from six directional domain transfers, three seeds, and five deployment regimes. Each world contains 2,000 target images, and all TTA actions run with execution batch size 32.

Matched IID/correlated worlds again contain the same selected image multiset and class quota; only order changes. The original stream-aware Z3 representation used a 200-sample evidence window. Final CPU-only reviewer checks then reconstruct the frozen evidence exactly and evaluate a compact 11-number stream summary at fixed windows 32, 64, 128, and 200 without rerunning any TTA method.

The 11 numbers measure variation and drift in SOURCE entropy, confidence, margin, predicted-class composition, and mean logits across successive windows. They use SOURCE outputs only. A nested-CV Ridge regression provides a deliberately simple linear baseline to test whether the result depends on random-forest nonlinearity.

\section{Results}
\label{sec:results}

\subsection{Same images, different deployment, different oracle action}
The matched experiments give the most direct test of the theory. On CIFAR-100-C, 157 of 270 matched IID/correlated pairs change oracle action even though each pair contains exactly the same selected images; 82 are strong flips. Global permutation-invariant evidence is essentially unchanged across each pair, so it cannot respond to an action reversal caused only by stream structure. On DomainNet-126, the oracle action changes in 33 of 36 same-image matched pairs, including 27 strong flips. Aligned SOURCE predictions and SOURCE error remain unchanged.

These are not synthetic toy outcomes: they arise from public benchmarks and modern TTA actions. They instantiate Proposition~\ref{prop:invariance}: an observation channel can discard a deployment variable that matters to the ranking of the actions.

\subsection{CIFAR-100-C: richer evidence sharply reduces oracle regret}
Table~\ref{tab:cifar-main} reports the frozen leave-one-corruption-out controlled evaluation. Z3 reduces mean regret to 0.312 pp. The final author-confirmed MORPHEUS MultiOutput baselines remain substantially farther from oracle: 1.170 pp for the text Neural-Collapse definition and 1.358 pp for the appendix definition. The entropy-only MORPHEUS variant has 4.135 pp regret. Paired bootstrap intervals and Holm-corrected tests comparing these MORPHEUS variants with Z3 are reported in Appendix~\ref{app:experiments}.

\begin{table}[t]
\centering
\small
\caption{CIFAR-100-C controlled LOCO selection. Lower regret is better; action accuracy is the fraction of worlds where the exact oracle action is selected. MORPHEUS uses the final author-confirmed MultiOutput RF implementation.}
\label{tab:cifar-main}
\begin{tabular}{lrr}
\toprule
Selector / evidence & Regret (pp) $\downarrow$ & Action acc. (\%) $\uparrow$\\
\midrule
SOURCE / KEEP & 4.980 & 50.2\\
MORPHEUS entropy & 4.135 & 53.0\\
MORPHEUS NC-text & 1.170 & 67.6\\
MORPHEUS NC-appendix & 1.358 & 66.5\\
Z1 (global) & 1.705 & 63.7\\
Z2 (richer global) & 1.247 & 65.0\\
\textbf{Z3 (stream-aware)} & \textbf{0.312} & \textbf{76.6}\\
Oracle & 0.000 & 100.0\\
\bottomrule
\end{tabular}
\end{table}

The matched strong-flip subset makes the mechanism even clearer. Z3 changes its selected action in 92.7\% of the 82 strong pairs and has only 0.093 pp mean pair regret. The author-confirmed MORPHEUS entropy, NC-text, and NC-appendix selectors change action in 0\% of these pairs and have 6.189, 2.873, and 2.921 pp pair regret, respectively. Because their evidence is global, the two reorderings look the same to them.

\subsection{DomainNet-126: the signal is small, simple, and execution-scale}
DomainNet independently reproduces the information-boundary effect under natural domain shifts. Table~\ref{tab:domainnet-main} shows the leave-one-transfer-out result. The original 2,990-dimensional Z3 RF has 0.034 pp regret and 93.3\% exact action accuracy. The predeclared 11-dimensional ORDER11 summary at the actual TTA execution scale $W=32$ gives the \emph{same} point result. A nested Ridge model on those 11 numbers remains strong at 0.061 pp regret and 92.2\% action accuracy. In contrast, global Z2 remains at 1.933 pp regret and 45.6\% action accuracy.

\begin{table}[t]
\centering
\small
\caption{DomainNet-126 LOTO selection over 90 worlds. The compact ORDER11 results use only 11 SOURCE-derived stream statistics.}
\label{tab:domainnet-main}
\begin{tabular}{lrr}
\toprule
Selector / evidence & Regret (pp) $\downarrow$ & Action acc. (\%) $\uparrow$\\
\midrule
SOURCE / KEEP & 1.933 & 45.6\\
MORPHEUS entropy & 4.886 & 46.7\\
Z2 RF (global) & 1.933 & 45.6\\
Z3 MultiOutput & 0.331 & 83.3\\
Z3 RF (2,990-D) & 0.034 & 93.3\\
\textbf{ORDER11 RF, $W=32$} & \textbf{0.034} & \textbf{93.3}\\
Ridge ORDER11, $W=32$ & 0.061 & 92.2\\
\bottomrule
\end{tabular}
\end{table}

On the 27 strong DomainNet matched flips, global Z2 changes action in 0/27 pairs and has 1.942 pp mean pair regret. Z3 and ORDER11-RF change action in all 27 with zero pair regret. Ridge changes action in 26/27 and has 0.044 pp pair regret. The conclusion therefore does not depend on thousands of features, a $W=200$ window, or a complicated nonlinear selector. Window, seed, PCA, transfer-holdout, and native-TTA checks are in Appendix~\ref{app:experiments}.

\subsection{No evidence representation is universally best}
The controlled matched-world experiment is deliberately designed so that order is decision-relevant. Under randomized stochastic CIFAR deployments, the large Z3 advantage disappears. When selectors are trained and evaluated within the stochastic deployment family, MORPHEUS NC-text obtains 0.210 pp mean regret, NC-appendix 0.293 pp, and Z3 0.338 pp. We keep this counter-result because it is important: richer stream evidence is not automatically superior. Its value depends on whether the deployment family contains ambiguities that the extra information resolves.

This is the empirical form of the paper's central claim. A selector should not be judged only by its prediction architecture; it should also be judged by whether its observation channel retains the information needed to distinguish deployments whose action rankings differ.

\section{Related Work and Positioning}
\label{sec:related}

\paragraph{TTA reliability and deployment effects.}
TTA methods such as Tent, EATA, SAR, DeYO, and ROID focus primarily on how to adapt reliably \citep{wang2021tent,niu2022eata,niu2023sar,lee2024deyo,marsden2024roid}. NOTE, TTN, and realistic TTA evaluations show that temporal correlation, small batches, class imbalance, and protocol choices can strongly affect adaptation \citep{gong2022note,lim2023ttn,zhao2023pitfalls}. Tempora further shows that TTA utility and method rankings can change over time \citep{sreeram2026tempora}. These findings motivate our question, but they do not characterize whether a specified pre-action evidence channel identifies the oracle intervention.

\paragraph{TTA performance estimation and method selection.}
AETTA estimates the accuracy of an adapted model without labels, while TTALine uses accuracy/agreement relationships after TTA \citep{lee2024aetta,kim2024ttaline}. MORPHEUS is our closest operational neighbor: it uses pre-adaptation entropy and Neural-Collapse geometry to predict post-adaptation performance and select among TTA procedures \citep{danilowski2026morpheus}. We do not claim to be the first to select a TTA method before adaptation. We ask the prior statistical question: \emph{when can any selector using a given pre-action observation channel determine the oracle action at all?}

Cygert et al. study realistic unsupervised TTA model selection, but their criteria are evaluated after candidate adaptations are run \citep{cygert2026realistic}. This is a richer information/computation budget than our pre-action setting, so we treat it as an adjacent comparison rather than forcing a misleading same-budget table. NEO uses latent re-centering as an actual TTA method \citep{murphy2026neo}; our \recent action is instead a one-dimensional theoretical device used to derive a finite-sample boundary.

\paragraph{Recent limits and underspecification in TTA.}
Zhou et al. study TTA learnability through recovery complexity: how quickly an adaptation process can return to low excess risk after a shift \citep{zhou2026learnability}. Shamsi et al. study underspecification of unsupervised TTA objectives and maintain multiple hypotheses \citep{shamsi2026multihypothesis}. These are close in spirit because they expose limits of unlabeled adaptation, but their objects differ from ours. We ask whether the identity of the best member of a fixed menu of data-dependent interventions is determined by a specified observation channel \emph{before} choosing an intervention.

\paragraph{Domain adaptation, transfer, and unlabeled performance estimation.}
Ben-David et al. give foundational impossibility results for domain adaptation, and Garg et al. show that unlabeled OOD performance estimation requires assumptions \citep{benDavid2010impossibility,garg2022atc}. Gulrajani and Hashimoto study identifiability of domain mappings; Kong et al. study partial identifiability of target-domain structure; more recent representation-based work identifies still richer target information under structural assumptions \citep{gulrajani2022identifiability,kong2022partial,ng2025representation}. These works target a domain map, target structure, or predictor-level object. Our target is deliberately downstream: the index of the best TTA intervention for a fixed source model.

Hanneke et al. analyze model selection and adaptation to unknown transfer relationships and oracle rates \citep{hanneke2023limits}. Dong et al. analyze the hardness of unsupervised domain adaptation through information about target labels and the attainable target predictor \citep{dong2026hardness}. Their settings are important theoretical neighbors, but neither directly gives our finite-batch KEEP/RECENTER boundary, our pre-action TTA observation-channel criterion, or the matched-stream invariance result.

\paragraph{Decision making under partial identification.}
The abstract principle that an optimal decision or policy ranking can sometimes be determined without identifying every payoff is well established in statistics and econometrics \citep{kasy2016partial,pu2021optimal,han2024dynamic,christensen2026optimal}. We therefore do \emph{not} present Action-ID, the two-point lower bound, or the TV inequality as new general decision theory. We use that machinery as a foundation and concentrate the contribution on TTA-specific information boundaries: data-dependent interventions, finite adaptation batches, shift-mechanism reversals, and observation channels that can erase or preserve deployment structure.

\section{Discussion and Limitations}
Our results expose a failure mode that is easy to miss when TTA selection is treated only as a prediction problem. A selector can fail because its regression model is weak, but it can also fail because the evidence it receives makes deployments with different action rankings indistinguishable. The second failure cannot be repaired by simply increasing selector capacity.

This matters for current TTA practice. Strong adaptation methods and increasingly sophisticated selection criteria do not remove the need to ask what information reaches the selector. On CIFAR-100-C and DomainNet-126, the same selected images can require different actions after only the stream organization changes. Global evidence can therefore be insufficient even when it contains hundreds of source-output statistics. Conversely, a small stream-aware summary can recover the missing decision signal when order is the relevant hidden variable.

The conclusion is also not that order-aware evidence is always better. Our stochastic experiment shows the opposite: when the deployment family changes, the large controlled advantage can disappear. This makes the claim stronger and more useful, not weaker. Reliable TTA selection requires matching the observation channel to the deployment variations that can change the action ranking.

There are clear limits. The theory uses a finite action menu and a simple one-dimensional Gaussian model. The Gaussian RECENTER action is a mechanism study, not a literal model of every modern TTA algorithm. The experiments use one main source-model family, and the DomainNet action boundary is simpler than CIFAR because correlated worlds strongly favor SOURCE. Broader architectures and action menus are natural future tests.

Finally, the translation/prior-shift construction is intentionally a coarse-evidence result. The two full target distributions differ; only the specified population-mean channel is shared. Richer observations may resolve the ambiguity. That is the central lesson: identifiability belongs to the pair \emph{(deployment family, observation channel)}. The practical design question is therefore not only ``which selector should we train?'' but also ``what information must that selector be allowed to see for the desired TTA decision to be possible?''

\section*{Reproducibility Statement}
The experiment notebooks freeze the target worlds, actions, random seeds, evidence definitions, selectors, and statistical tests. Full mathematical proofs are included in Appendix~\ref{app:proofs}, and the complete experimental audit is summarized in Appendix~\ref{app:experiments}. The final anonymous submission will include the code and frozen result artifacts required to reproduce the reported tables.

\section*{AI Use Statement}
Generative AI tools were used during the research workflow for code debugging, organization, mathematical checking, and language editing. The authors remain responsible for every theorem assumption, proof, experiment, numerical claim, citation, and manuscript statement.

\section*{Acknowledgements}
This research is supported by the Ministry of Education, Singapore, under its funding for the Research Centre of Excellence award to the Institute for Digital Molecular Analytics \& Science (IDMxS), NTU. Project No. EDUNC-33-18-279-V12-IDMxS.
\bibliographystyle{iclr2027_conference}
\bibliography{references}

\appendix
\section{Appendix}
\label{app:proofs}

This appendix gives the formal steps. Before each proof we first state the idea in plain language.

\subsection{Proof of Theorem~\ref{thm:exact}} \label{proof1}
\paragraph{Idea.}
The selector receives statistically identical evidence in both worlds. Therefore it must use the same action probabilities in both. Since the two correct actions are different, the probability mass assigned to ``correct in world 0'' and ``correct in world 1'' cannot add to more than one.

Let $p_j(z)$ be the selector's conditional probability of outputting $a_j^*$ after observing $z$, including any internal randomization. Because the two evidence laws are the same, call the common law $Q$. Then
\[
\Pbb_{\theta_0}(s(Z_n)=a_0^*)+\Pbb_{\theta_1}(s(Z_n)=a_1^*)
=\int\!\bigl[p_0(z)+p_1(z)\bigr]dQ(z)\leq1,
\]
because $a_0^*\neq a_1^*$ implies $p_0(z)+p_1(z)\leq1$ for every $z$. Under an equal prior, average correctness is therefore at most $1/2$, so average action error is at least $1/2$. If every wrong action has excess risk at least $\Delta$, regret is at least $\Delta$ whenever the selector is wrong, giving average regret at least $\Delta/2$. \hfill$\square$

\paragraph{Explicit Discrete Two-World Construction.}
To instantiate Theorem~\ref{thm:exact} with an explicit discrete example, let $\mathcal{X} = \{x_1, x_2, x_3\}$ with identical target marginal distribution $P(X) = (0.45, 0.25, 0.30)$ in both worlds. Consider two candidate classifiers:
\begin{align*}
f_{\text{KEEP}}(x_1) &= 1, \quad f_{\text{KEEP}}(x_2) = 0, \quad f_{\text{KEEP}}(x_3) = 0, \\
f_{\text{ADAPT}}(x_1) &= 0, \quad f_{\text{ADAPT}}(x_2) = 1, \quad f_{\text{ADAPT}}(x_3) = 0.
\end{align*}
Let $\eta_\theta(x) = \mathbb{P}_\theta(Y = 1 \mid X = x)$ define the target posterior label distribution:
\begin{align*}
\text{World A: } \quad \eta_A &= (0.250, 0.675, 0.1875), \\
\text{World B: } \quad \eta_B &= (0.750, 0.315, 0.0625).
\end{align*}
Under 0-1 loss $R_\theta(f) = \sum_{i=1}^3 P(X = x_i) \mathbb{P}_\theta(f(x_i) \neq Y \mid X = x_i)$, the exact prospective risks evaluate to:
\begin{align*}
R_A(\text{KEEP}) &= 0.45(1 - 0.250) + 0.25(0.675) + 0.30(0.1875) = 0.5625, \\
R_A(\text{ADAPT}) &= 0.45(0.250) + 0.25(1 - 0.675) + 0.30(0.1875) = 0.2500, \\
R_B(\text{KEEP}) &= 0.45(1 - 0.750) + 0.25(0.315) + 0.30(0.0625) = 0.2100, \\
R_B(\text{ADAPT}) &= 0.45(0.750) + 0.25(1 - 0.315) + 0.30(0.0625) = 0.5275.
\end{align*}
Thus, $\mathcal{A}^*(\text{World A}) = \{\text{ADAPT}\}$ while $\mathcal{A}^*(\text{World B}) = \{\text{KEEP}\}$. Since the observation channel $Z_n \sim P(X)^{\otimes n}$ is statistically identical in both worlds, any pre-adaptation selector incurs an average action error of at least $1/2$ and an expected regret of at least $\Delta / 2 = 0.15875$.

\subsection{Proof of Corollary~\ref{cor:tv}} \label{proof2}
\paragraph{Idea.}
When the two evidence laws are not identical, the selector can partly tell them apart. The best possible two-way discrimination advantage is controlled by total variation. Whatever ambiguity remains becomes an unavoidable action-selection error.

Let $A_0$ be the event that the selector outputs $a_0^*$. Since $a_1^*\neq a_0^*$,
\[
\Pbb_0(\text{correct})+\Pbb_1(\text{correct})
\leq Q_0(A_0)+Q_1(A_0^c).
\]
The right side is
\[
1+Q_0(A_0)-Q_1(A_0)
\leq1+\TV(Q_0,Q_1).
\]
Average correctness is therefore at most $(1+\TV)/2$, so average error is at least $(1-\TV)/2$. Multiplying by the minimum wrong-action cost $\Delta$ gives the regret bound. \hfill$\square$

\subsection{Proof of Theorem~\ref{thm:positive}} \label{proof3}
\paragraph{Idea.}
If all worlds compatible with the evidence share one optimal action, choose it. If they do not share any optimal action, no single decision based only on that same evidence law can be optimal in every compatible world.

If
$a\in\bigcap_{\theta\in\ThetaSet(Q)}\A^*(\theta)$,
then the decision rule $d(Q)=a$ has zero oracle regret in every compatible world. Conversely, if the intersection is empty, every action fails to be optimal in at least one compatible world. Hence no action-valued function of $Q$ can have zero oracle regret simultaneously across the whole compatible set. \hfill$\square$

\subsection{Proof of Proposition~\ref{prop:invariance}} \label{proof4}
\paragraph{Idea.}
An invariant observation maps a batch and its transformed version to exactly the same value. Therefore the transformed worlds have the same evidence law. If their best actions differ, Theorem~\ref{thm:exact} applies.

Let the raw batch in world $\theta$ be $B$. The transformed world uses the pushforward distribution of $gB$. Since $\phi_n(gB)=\phi_n(B)$ pointwise, both worlds induce the same evidence law. Different unique oracle actions then imply non-identifiability by Theorem~\ref{thm:exact}. \hfill$\square$

\subsection{Proof of Theorem~\ref{thm:gaussian_boundary}} \label{proof5}
\paragraph{Step 1: write the KEEP risk as one function.}
Define
\[
g(t)=\frac12\left[
\Phi\!\left(-\frac{\mu+t}{\sigma}\right)
+\Phi\!\left(\frac{t-\mu}{\sigma}\right)
\right].
\]
This is exactly the error of the fixed source threshold when the target translation is $t$, so $R_K(\delta)=g(\delta)$.

The function is even: a left shift and an equally large right shift hurt the symmetric source classifier by the same amount. For $t>0$,
\[
g'(t)=\frac{1}{2\sigma}\left[
\varphi\!\left(\frac{t-\mu}{\sigma}\right)
-\varphi\!\left(\frac{t+\mu}{\sigma}\right)
\right]>0.
\]
The inequality holds because $|t-\mu|<t+\mu$ and the standard normal density decreases as $|x|$ moves away from zero. Therefore $g$ strictly increases with $|t|$. Also
\[
g(0)=\Phi(-\mu/\sigma),
\qquad
\lim_{t\to\infty}g(t)=1/2.
\]

\paragraph{Step 2: understand what RECENTER leaves behind.}
The estimate is
\[
\widehat\delta=\bar X_n
=\delta+\frac1n\sum_{i=1}^n(\mu Y_i+\varepsilon_i).
\]
Define the estimation error
\[
E_n=\widehat\delta-\delta.
\]
After moving the threshold by $\widehat\delta$, the true translation $\delta$ is cancelled and only the random estimation error remains. Because the evaluation point is independent of the adaptation batch,
\[
R_R(n)=\Ebb[g(E_n)].
\]
For every finite $n$, $E_n$ is non-degenerate, so $E_n\neq0$ almost surely. Since $g(t)>g(0)$ whenever $t\neq0$,
\[
R_R(n)>g(0).
\]
On the other hand $g(t)<1/2$ for every finite $t$, hence $R_R(n)<1/2$.

Because $g$ is continuous and strictly increasing on $t>0$, there is exactly one positive value $\delta_c(n)$ with
\[
g(\delta_c(n))=R_R(n).
\]
Below that value KEEP has lower risk; above it RECENTER has lower risk. This proves the existence, uniqueness, and ordering parts of the theorem.

\paragraph{Step 3: derive the exact finite-$n$ RECENTER risk.}
Condition on
\[
S_n=\sum_{i=1}^nY_i=s.
\]
Then
\[
E_n\mid S_n=s
\sim\Normal\!\left(\frac{\mu s}{n},\frac{\sigma^2}{n}\right).
\]
For a Gaussian random variable $Z\sim\Normal(m,\tau^2)$, the identity
\[
\Ebb\left[\Phi\!\left(\frac{Z-a}{\sigma}\right)\right]
=\Phi\!\left(\frac{m-a}{\sqrt{\sigma^2+\tau^2}}\right)
\]
follows by introducing an independent standard normal variable and combining two Gaussian noises. Averaging over the binomial values $S_n=2k-n$ gives Eq.~\eqref{eq:recenter-risk}. Symmetry of $E_n$ makes the two class-error terms equal after expectation.

\paragraph{Step 4: find the large-$n$ scale of the crossing.}
Near zero, $g$ is smooth and even, so the first non-zero term is quadratic:
\[
g(t)=g(0)+c_2t^2+O(t^4),
\qquad
c_2=\frac{\mu\,\varphi(\mu/\sigma)}{2\sigma^3}>0.
\]
The estimation error is a sample mean of $\mu Y+\varepsilon$, whose variance is
\[
v=\mu^2+\sigma^2.
\]
Therefore
\[
\Ebb[E_n^2]=\frac{v}{n},
\qquad
\Ebb[E_n^4]=O(n^{-2}).
\]
Taking expectation in the Taylor expansion gives
\[
R_R(n)=g(0)+c_2\frac{v}{n}+O(n^{-2}).
\]
At the crossing, $g(\delta_c(n))=R_R(n)$. Since $\delta_c(n)\to0$,
\[
c_2\delta_c(n)^2+O(\delta_c(n)^4)
=c_2\frac{v}{n}+O(n^{-2}),
\]
so
\[
\delta_c(n)^2=\frac{v}{n}+O(n^{-2}).
\]
Taking the positive square root yields
\[
\delta_c(n)=
\frac{\sqrt{\mu^2+\sigma^2}}{\sqrt n}
+O(n^{-3/2}).
\]
This proves the theorem. \hfill$\square$

\subsection{Proof of Proposition~\ref{prop:prior}} \label{proof6}
\paragraph{Idea.}
A change in class proportions moves the mixture mean and the Bayes decision boundary in opposite directions. Mean recentering follows the mixture mean, so it moves away from the Bayes correction.

Under pure prior shift, $\Pbb(Y=+1)=\pi$. For threshold $t$,
\[
R_\pi(t)=
\pi\Phi\!\left(\frac{t-\mu}{\sigma}\right)
+(1-\pi)\Phi\!\left(\frac{-t-\mu}{\sigma}\right).
\]
The Bayes-optimal threshold is
\[
t_B=\frac{\sigma^2}{2\mu}\log\frac{1-\pi}{\pi}.
\]
If $\pi>1/2$, then $t_B<0$, while the population mean
\[
m=\mu(2\pi-1)>0.
\]
The derivative of $R_\pi(t)$ changes sign only at $t_B$ and is positive for $t>t_B$. Therefore $R_\pi(t)$ is strictly increasing from $0$ to $m$, giving
\[
R_\pi(m)>R_\pi(0).
\]
Thus moving the threshold to the target mean is worse than keeping it at zero. The case $\pi<1/2$ follows by symmetry. \hfill$\square$

\subsection{Proof of Corollary~\ref{cor:mean}} \label{proof7}
\paragraph{Idea.}
We build two worlds with the same reported mean $m$. In one world the mean moved because the whole distribution translated, so recentering fixes the problem. In the other world the mean moved only because the class proportions changed, so recentering moves the threshold the wrong way.

In World C, balanced translation by $m$ gives population mean $m$. Population RECENTER uses the true mean and returns the source risk $g(0)$, while KEEP has $g(m)>g(0)$.

In World P, choose
\[
\pi=\frac{1+m/\mu}{2},
\]
which is valid for $|m|<\mu$. Then the target mean is also
\[
\mu(2\pi-1)=m.
\]
But Proposition~\ref{prop:prior} gives
\[
R_\pi(0)<R_\pi(m),
\]
so KEEP is better than population RECENTER. Thus the same population mean corresponds to opposite best actions.

For the finite-batch RECENTER action, the opposite-action conclusion also holds for all sufficiently large $n$. In World C, Theorem~\ref{thm:gaussian_boundary} gives $\delta_c(n)\to0$, so a fixed nonzero $m$ eventually lies above the boundary. In World P, the sample mean converges almost surely to $m$, and the bounded fresh-point classification loss converges to the strictly worse population-RECENTER risk. \hfill$\square$

\section{Additional Experimental Details}
\label{app:experiments}

This appendix records the frozen reviewer checks behind the concise main-text tables. No result below is chosen by selecting the most favorable random seed or window after seeing the outcome.

\subsection{CIFAR-100-C: complete controlled study}
The controlled benchmark has 675 worlds: 15 corruption types $\times$ three severities $\times$ five deployment regimes $\times$ three seeds. The five actions are SOURCE, DeYO, Tent, ROID, and test-time normalization. The oracle counts are SOURCE 339, ROID 277, Tent 34, and DeYO 25 worlds. The matched IID/correlated experiment contains 270 pairs with exactly the same selected images; 157/270 change oracle action and 82 are strong flips.

The frozen Z3 selector has 0.312 pp mean oracle regret in leave-one-corruption-out evaluation. The final author-confirmed MORPHEUS MultiOutput-RF baselines are 4.135 pp for entropy, 1.170 pp for the text Neural-Collapse definition, and 1.358 pp for the appendix Neural-Collapse definition. Relative to Z3, the mean paired regret differences are approximately +3.823, +0.857, and +1.046 pp; the corresponding transfer-level tests remain significant after Holm correction ($p\approx3.66\times10^{-4}$).

On the 82 strong same-image flips, frozen Z3 changes its selected action in 92.7\% of pairs with 0.093 pp mean pair regret. The author-confirmed MORPHEUS entropy, NC-text, and NC-appendix selectors change action in 0\% of these pairs and have mean pair regrets 6.189, 2.873, and 2.921 pp.

\paragraph{Reviewer checks.}
The frozen audit includes exact/tie-aware oracle metrics; five independent RF seeds; feature-dimension control; paired bootstrap confidence intervals; paired significance tests; conventional untouched CIFAR-100-C sanity checks; and a batch-size frontier. Compressing Z3 to five dimensions is worse than MORPHEUS, while increasing retained information improves performance and roughly 64 dimensions becomes competitive/better. This is useful evidence that the result depends on preserving decision-relevant information rather than simply naming one representation ``better.''

The batch-size frontier also confirms that the oracle action itself can move as execution scale changes. Relative to the $B=200$ reference, action-switch rates are 50.0\% at $B=20$, 31.25\% at $B=50$, 16.67\% at $B=100$, and 29.17\% at $B=500$ in the audited subset.

\subsection{CIFAR-100-C: stochastic counter-result}
The controlled stream construction is not the only deployment family we test. Under randomized stochastic deployments, the large Z3 advantage disappears. When selectors are trained and evaluated within the stochastic family, mean regret is 0.210 pp for MORPHEUS NC-text, 0.293 pp for NC-appendix, and 0.338 pp for Z3. We retain this result because it rejects a universal-superiority interpretation of Z3 and supports the observation-channel/deployment-family framing.

\subsection{Cygert et al.: stronger information budget}
The Cygert et al. native source/model/data sanity reproduces their published CIFAR-100-C SOURCE behavior closely. Their released evaluation pipeline does not expose every ENT/CON/SND statistic cleanly in our reproducible run, and some released runner paths are inconsistent for candidate methods. We therefore do not alter their scientific source code to force a same-table result. More importantly, their model-selection information is available after candidate adaptations have been executed, whereas our main setting chooses from pre-action evidence. We report this as a stronger-information-budget neighboring protocol rather than a direct same-budget baseline.

\subsection{DomainNet-126: frozen natural-domain study}
The study uses six directional transfers: real$\to$clipart, real$\to$painting, real$\to$sketch, clipart$\to$sketch, clipart$\to$real, and clipart$\to$painting. With three seeds and five regimes this gives 90 worlds. Each world uses 2,000 images without replacement. All five TTA actions run at execution batch size 32.

The oracle action is SOURCE in 41 worlds, ROID in 48, and DeYO in one. All 36 correlated worlds prefer SOURCE, while the 54 IID worlds are mostly ROID. This makes DomainNet a deliberately clean mechanism confirmation rather than the most difficult possible selector benchmark.

Matched IID/correlated worlds share the same selected images and class quota. The oracle changes in 33/36 pairs and 27 are strong flips. Final numerical checks align samples by original dataset index and verify SOURCE-prediction and SOURCE-error consistency. Canonical permutation invariance is checked on the same saved tensor rather than by demanding bitwise equality between separate GPU forwards.

\paragraph{Main and compact evidence.}
Across all 90 worlds, global Z2 has 1.933 pp mean regret and 45.6\% exact action accuracy. Original Z3 RF has 0.034 pp regret and 93.3\% action accuracy. A CPU-only audit reconstructs the original $W=200$ Z2/Z3 evidence bitwise exactly, then evaluates the already-predeclared 11-number ORDER11 summary at windows 32, 64, 128, and 200. ORDER11-RF gives 0.034 pp/93.3\% at $W=32$, 0.106 pp/91.1\% at $W=64$, 0.034 pp/93.3\% at $W=128$, and 0.034 pp/93.3\% at $W=200$. Thus the execution-scale $W=32$ result is identical to the original $W=200$ point prediction.

Adding global Z2 to ORDER11 does not improve the $W=32$, 128, or 200 point metrics, which supports the interpretation that stream structure is the decisive signal in this construction. On the 27 strong matched flips, ORDER11 changes action in all 27 with zero mean pair regret at every tested window.

\paragraph{Linear-model check.}
Nested Ridge regression chooses regularization using only the training transfers. With ORDER11 at $W=32$, Ridge has 0.061 pp mean regret, 92.2\% exact action accuracy, 96.7\% within 0.5 pp of oracle, and no 5-pp catastrophic cases. On the 27 strong matched flips it changes action in 26/27 pairs and has 0.044 pp mean pair regret. The same Ridge architecture on global Z2 remains at 1.933 pp regret, 45.6\% action accuracy, and 0/27 action changes.

\paragraph{Robustness and generalization checks.}
Five RF seeds give identical main Z3 metrics. Fold-only PCA at 5, 16, 32, and 64 dimensions gives the same 0.034 pp/93.3\% point result on DomainNet, showing that the mechanism is low-dimensional there. Leave-one-target-domain-out regret is 0.180 pp for clipart, 0 for painting, 0.027 pp for real, and 0 for sketch. Leave-one-source-domain-out is harder for the clipart-source family (1.171 pp regret) but still improves substantially over SOURCE (2.522 pp). The real-source conventional native sequence also behaves normally: ROID improves over SOURCE on the standard domain sequence, confirming that the severe correlated-world adaptation failures are not simply a broken implementation.

\subsection{What is deliberately not claimed}
The studies do not claim that Z3 or ORDER11 is a universally best selector. They also do not claim that every TTA action reversal is caused by order. Their purpose is narrower: to show concrete public-benchmark deployment families where a restricted observation channel collapses worlds with different oracle actions, and to show that adding the missing kind of information can recover the decision in those families.

\end{document}